\documentclass[runningheads]{llncs}
\usepackage{graphicx}
\usepackage{subcaption}
\usepackage{amsfonts}

\begin{document}
\title{Optimizing the Wisdom of the Crowd: \\ Inference, Learning, and Teaching}
%
%\titlerunning{Abbreviated paper title}
% If the paper title is too long for the running head, you can set
% an abbreviated paper title here
%
\author{Yao Zhou \and Jingrui He}
\authorrunning{Yao Zhou et al.}
% First names are abbreviated in the running head.
% If there are more than two authors, 'et al.' is used.
%
\institute{Arizona State University, Tempe AZ 85282, USA \\  \email{\{yzhou174, jingrui.he\}@asu.edu}}
\maketitle              % typeset the header of the contribution
%
% \begin{abstract}
% The abstract should briefly summarize the contents of the paper in
% 15--250 words.

% \keywords{First keyword  \and Second keyword \and Another keyword.}
% \end{abstract}
%
%
%
\vspace{-5mm}
\setlength{\textfloatsep}{2pt}% Remove \textfloatsep
\section{Introduction}
\vspace{-3mm}
\subsection{Motivation and Problem Description.}
\begin{figure}[t]
    \centering
    \begin{subfigure}[b]{0.53\textwidth}
        \includegraphics[width=\textwidth]{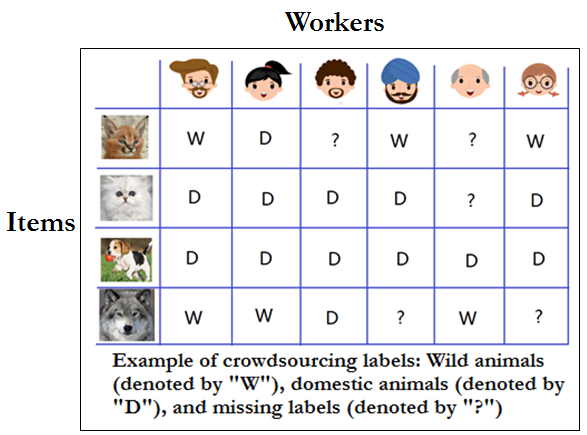}
        \caption{An illustration of crowdsourcing on categorization.}
        \label{crowdsourcing}
    \end{subfigure}
    ~ %add desired spacing between images, e. g. ~, \quad, \qquad, \hfill etc. 
      %(or a blank line to force the subfigure onto a new line)
    \begin{subfigure}[b]{0.44\textwidth}
        \includegraphics[width=\textwidth]{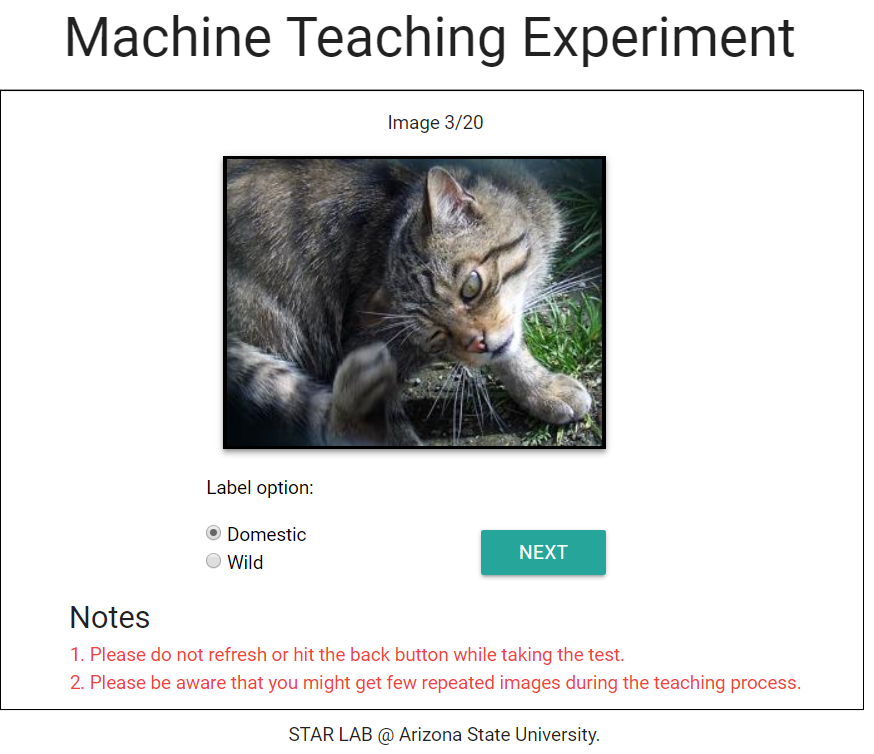}
        \caption{One teaching iteration of JEDI framework}
        \label{JEDI}
    \end{subfigure}
\end{figure}

\vspace{-2mm}
The unprecedented demand for large amount of data has catalyzed the trend of combining human insights with machine learning techniques, which facilitate the use of crowdsourcing to enlist label information both effectively and efficiently. The classic work on crowdsourcing mainly focuses on the label inference problem under the categorization setting, as shown in Figure. \ref{crowdsourcing}. However, inferring the true label requires sophisticated aggregation models that usually can only perform well under certain assumptions. Meanwhile, no matter how complicated the aggregation model is, the true model that generated the crowd labels remains unknown. Therefore, the label inference problem can never infer the ground truth perfectly.

% \subsection{Problem Description.}
Based on the fact that the crowdsourcing labels are abundant and utilizing aggregation will lose such kind of rich annotation information (e.g., which worker provided which labels), we believe that it is critical to take the diverse labeling abilities of the crowdsourcing workers as well as their correlations into consideration. To address the above challenge, we intend to tackle three research problems, namely {\bf inference}, {\bf learning}, and {\bf teaching}. The detailed descriptions of these tasks are list as below:
\vspace{-1mm}
\begin{itemize}
    \item \textbf{Crowdsourcing Label Inference:} In this task, our aim is to infer the true labels of all items (e.g., images or documents) by providing the crowdsourced label matrix $M \in \mathbb{R}^{N \times N_w}$, where $N$ is the number of items and $N_w$ is the number of non-expert workers, as input. The collected crowd label matrix $M$ usually contains noisy (i.e., incorrect) labels and missing entries.
    \item \textbf{Heterogeneous Learning with Crowdsourcing:} In this task, we propose to perform heterogeneous model learning (e.g., multi-task learning, multi-view learning) by exploring and utilizing the intrinsic structural correlation between the collected labels of these crowdsourcing workers. In this setting, we bypass the conventional two-stage learning procedure which requires label inference and model learning as two separate steps.
    \item \textbf{Adaptive Crowd Teaching:} In this task, we utilize crowdsourcing by supervising the workers to label in form of teaching because human beings are extremely good at learning a specific target concept and transfer the learned concepts to solve similar problems. With a properly designed teacher which can teach the crowdsourcing workers to label with a few representative examples and estimate the learning progress of each worker, the collected crowdsourcing labels could have higher quality and less missing answers.
\end{itemize}

\vspace{-7mm}
\subsection{Background and Related Work.}
\vspace{-2mm}
The classic label inference problem has been extensively studied in the machine learning community. The classic Dawid-Skene EM model \cite{Dawid1979maximumlikelihood} treats the worker ability confusion matrix as a latent vector and it is often referred as the two-coin model. Raykar’s EM algorithm \cite{Vikas} impose a beta prior over the worker confusion matrix and it jointly learns the classifier and true labels together. The minimax entropy model proposed by Zhou et al. \cite{MMCE} is able to further infer the true labels, item difficulty, and worker ability jointly. In contrast, our work \cite{TAC,MultiC2,M2VW} can leverage the structural information from multiple types of data heterogeneity, e.g. task heterogeneity, view heterogeneity, and oracle heterogeneity, by applying the rank minimization principles on the tensor represented data. This principle is generic to be utilized on both the label inference problem as well as the heterogeneous learning problem. Furthermore, we also develop a framework \cite{JEDI} for personalized crowd teaching that considers both the teaching usefulness and teaching diversity.

\vspace{-5mm}
\section{Our work}
\vspace{-3.5mm}
\subsection{Proposed Solutions.}
\vspace{-1.5mm}
In this section, we present our solutions for these problems we mentioned above:
\vspace{-8mm}
\subsubsection{\small Tensor Augmentation and Completion (TAC):} TAC \cite{TAC} is a novel structural approach which uses the tensor representation for the labeled data, augments it with a ground truth layer, and then utilized two distinct regularization terms to estimate the ground truth layer via low-rank tensor completion method. 
\vspace{-5mm}
\subsubsection{\small Multi-task Classifications using Crowd Labels (MultiC$^2$):} MultiC$^2$ \cite{MultiC2} is an optimization framework, which uses a weight tensor to represent the crowdsourcing workers' behavior across multiple intrinsically correlated tasks. We also propose an iterative algorithm to solve the optimization problem and construct a worker ensemble based on the estimated tensor, whose decisions will be weighted using a set of entropy weights. 
\vspace{-5mm}
\subsubsection{\small Multi-view Multi-worker Heterogeneous Learning (M2VW):}
We formulate the multi-view multi-worker classification problem \cite{M2VW} as a regularized optimization problem which imposes the view consistency and the worker consensus principles through the low-rank prediction tensor regularization term.
\vspace{-5mm}
\subsubsection{\small Adjustable Exponential Decay Memory Interactive Crowd Teaching (JEDI):} We propose an adaptive teaching framework named JEDI \cite{JEDI} to construct the personalized optimal teaching set for the crowdsourcing workers who are assumed have exponentially decayed memory. Furthermore, the trade-off between the teaching usefulness and the teaching diversity is introduced in the framework to ensure learning comprehensiveness. 
\vspace{-4mm}
\subsection{Data Sets and Evaluation}
\vspace{-2mm}
We have utilized a wide variety of crowdsourcing data set for evaluations. (1). The classic data set, e.g., RTE, Temp, Web, Dog, Spam, and Age, which only have the crowd label matrix as the input are used for the evaluation of label inference problems. (2). The 20 Newsgroups and Animal Breed data set are used for the evaluations of heterogeneous learning problems as well as the adaptive crowd teaching problem. We use Accuracy and F1-score with multiple random splits and runs to evaluate classification performance. The efficiency of these algorithms are evaluated by running time, scalability, and convergence speed. The teaching effectiveness is evaluated by teaching gain for the real crowdsourcing workers.

\vspace{-5mm}
\section{Future work}
\vspace{-3mm}
Our future direction aims to dig deeper in both directions of crowd learning \cite{LiuL17} as well as crowd teaching. From the learning perspective, we would like to explore the performance of using embedded features combined with the crowdsourced labels. Another interesting direction would be self-annotation inspection of noisy labels by utilizing the influential function estimation \cite{influencefunction}. From the teaching perspective, we would like to provide the corresponding explanations (e.g., area of interest on images or phrases of documents) along with each teaching example.

\vspace{-4mm}
\bibliographystyle{splncs04}
\bibliography{sbp-doctor-consortium.bib}

\end{document}